# Three-Dimensional Retinal Microvasculature Restoration in OCT Angiography

Yukun Guo[1,2], Min Gao[1,2], Tristan T. Hormel[1], Steven T. Bailey[1], Thomas S. Hwang[1], Yali Jia[1,2,*]

[1]*Casey Eye Institute, Oregon Health & Science University, Portland, OR 97239, USA*
[2]*Department of Biomedical Engineering, Oregon Health & Science University, Portland, OR 97239, USA*
*[jiaya@ohsu.edu](mailto:jiaya@ohsu.edu)*

**Abstract:** Optical coherence tomographic angiography (OCTA) is a powerful technique for imaging retinal microvasculature. However, acquiring reliable quantification of retinal blood flow and areas of retinal nonperfusion is challenging because of imaging artifacts. Existing methods primarily focus on noise suppression, projection artifacts removal, or signal enhancement to improve image quality of OCTA on cross-sectional or two-dimensional (2D) *en face* projections, while neglecting the intrinsic three-dimensional vascular architecture. In this study, we propose a deep learning–based algorithm for restoring capillary anatomical vasculature from a single OCTA volume. The network consists of an EfficientNet-B5 encoder and a decoder incorporating concurrent spatial and channel squeeze-and-excitation modules, connected via skip connections to preserve spatial resolution. Three adjacent B-frames are used as input to predict the restored middle B-frame. We evaluated the performance of the model using the peak signal-to-noise ratio (PSNR) and structural similarity index measure (SSIM) against ground truth generated from averaging of multiple scans. The results show the proposed model significantly (both, $p<0.001$) improved image quality compare with original single OCTA volume, with PSNR of 26.16 ± 1.26 vs. 22.23 ± 0.78, and SSIM of 0.91 ± 0.02 vs. 0.72 ± 0.03. The proposed model also significantly ($p<0.001$) improved the microvascular fidelity, measured by the Dice coefficients overlap between module output and ground truth, in both 2D and 3D by at least 3.8/51.2% (2D/3D) in several different vascular slabs.

## Introduction

Optical coherence tomographic angiography (OCTA) can provide non-invasive, high-speed, and 3D imaging of the retinal microvasculature. Based on high-speed OCT, the signal in OCTA is acquired by computing the difference between two repeated OCT scans at the same raster position using an OCTA algorithm like optical microangiography (OMAG) [1], split-spectrum amplitude-decorrelation angiography (SSADA)[2], or Phase-stabilized complex-decorrelation angiography (PSCD) [3]. A major challenge in OCTA is the presence of artifacts, which can be broadly categorized into four types. First, speckle and background noise arise from interference patterns generated by the scattering of coherent laser light within microscopic tissue structures. Second, signal attenuation artifacts are typically caused by shadowing or defocus, leading to reduced signal intensity in deeper regions. Third, projection artifacts, where OCTA signals of superficial vessels is projected into the deeper layers, are inherent to the OCT imaging mechanism. Finally, motion artifacts result from eye movements or blinking; these can be substantially mitigated through eye-tracking and system-embedded registration algorithms integrated into modern OCTA systems. All these artifacts affect the visualization and downstream image analysis and quantification [4]. In one of our previous studies, we evaluated the impact of reduced signal strength on the quantification in OCTA[5]. In another study, we examined the influence of projection artifacts on macular neovascularization (MNV) segmentation [6]. Additionally, our prior work on nonperfusion area (NPA) segmentation accounted for signal attenuation artifacts in the quantification[7–10].

Several studies applied deep learning-based methods on OCTA image for suppressing noise on the OCTA. Jiang et al. applied the N2V2 framework, which is a self-supervised deep learning framework, to suppress noise on en face OCTA image[11]. The results show N2V2 framework can improve the OCTA image quality in most cases, however, the method can generate false vasculature from strong noise. Xu et al. proposed another self-supervised learning-based method for removing motion artifacts and noise on en face OCTA images [12]. Although this method reported high performance on the artifact removal, it is limited to 2D OCTA images. Projection artifacts in OCTA also affect the vasculature quantification in deep layer of retina. There are other methods based on traditional image processing techniques for noise reduction in OCTA [13–16]. Those methods, however, are not robust to the image quality due to their reliance on hand-crafted features and fixed assumptions, which limit their ability to generalize across varying noise levels, signal strengths, and imaging artifacts. To address projection artifacts, Zhang et al. introduced a slab-based subtraction strategy to suppress projection artifacts to improve the accuracy of CNV visualization[18]. However, because the method relies on slab-based subtraction, it may inadvertently attenuate true flow signals in deeper layers and lacks voxel-wise application preventing application to remove projection artifact on cross-sectional images. To mitigate the issue of signal attenuation with slab subtraction methods, our group developed projection-resolution OCTA (PR-OCTA) algorithm [17] that separates true in situ flow from projected signals. However, this algorithm is based on a simplified linear model that does not fully capture complex flow and reflectance interactions. Our group further improved projection artifact removal with an artificial intelligence-assisted projection removal method (aiPR-OCTA) with high accuracy on projection artifact removal [19] by incorporating structural information. Another work in our group further improved OCTA quality using deep learning-based *en face* reconstruction [20,21]; however, operating on 2D projections required additional post-processing for projection artifacts removal.

Despite these advances, existing approaches still exhibit notable limitations. These deep learning–based methods are designed for 2D *en face* OCTA images, neglecting the intrinsic 3D vascular architecture and limiting their ability to address artifacts that vary across depth. In addition, self-supervised frameworks, while effectively suppressing noise, may hallucinate false vasculature from noisy scans. Traditional image processing techniques, on the other hand, depend heavily on hand-crafted features and fixed assumptions, making them sensitive to variations in image quality and less adaptable to diverse imaging conditions. Projection artifact removal techniques, including slab-based subtraction and simplified modeling approaches, often fail to fully capture the complex interactions between flow and reflectance signals, and may suppress true vascular signals in deeper retinal layers. These limitations highlight the need for a more robust, volumetric approach that can simultaneously address multiple types of artifacts while preserving the integrity of the underlying microvascular structures. In this study, we developed a deep learning model to construct artifact-free OCTA volumes, capable of addressing spackle noise, background noise, projection artifacts, signal attenuation artifacts, and motion artifacts simultaneously while preserving true flow signals without hallucination.

## Methods

### *Data acquisition*

Volumetric OCTA scans were acquired from a 3 × 3 mm central macular region using a commercial OCT system operating at 120 kHz (SOLIX; Optovue/Visionix, Inc.). Two consecutive B-scans were acquired at each of the 400 raster scan positions, with each B-scan consisting of 400 A-lines. OCT volumes were then generated by averaging the paired B-scans at each location. Corresponding OCTA volumes were simultaneously generated by computing the intrinsic motion signal between consecutive B-scans using the split-spectrum amplitude-decorrelation angiography (SSADA) algorithm[2]. This study was approved by the Institutional Review Board of Oregon Health & Science University (Portland, OR) and was conducted in accordance with the Declaration of Helsinki.

### *Study Inclusion Criteria*

A total of 1,072 OCTA scans from one eye of 176 participants (32 healthy control, 55 age-related macular degeneration (AMD), and 89 diabetics retinopathy (DR)) were acquired, with 6–9 repeated scans per eye used to generate ground-truth volumes. All participants were diagnosed using the standard clinical criteria, ETDRS and AREDS, for AMD and DR. The scans with low signal quality (signal strength index < 50) were excluded, as well as eyes with fewer than six scans. The final dataset comprised 150 ground-truth OCTA volumes, with corresponding original single OCTA volumes randomly selected from the repeated sets. An additional dataset of 71 eyes with two repeated scans from a vision-threatening DR study was used to assess the model's repeatability.

***Deep learning model design***

The proposed network adopts an encoder–decoder architecture [Fig. 1] with an EfficientNet-B5 backbone[22] in the encoder and a decoder integrated with spatial and channel squeeze-and-excitation (scSE) modules[23], connected through skip connections to preserve resolution. The model processes one B-frame at a time while incorporating its two neighboring B-frames as contextual input. Specifically, three consecutive B-scans are stacked along the channel dimension to form a $488 \times 400 \times 3$ input, and the network predicts the restored middle B-frame ($488 \times 400 \times 1$), exploiting inter-slice structural continuity to suppress noise and projection artifacts. The full volume is reconstructed by sliding this three-frame window across the dataset with circular padding at the boundaries. The encoder follows the EfficientNet-B5 implementation, consisting of a convolutional stem and five downsampling stages built from stride-2 residual shortcut blocks and residual identity blocks, progressively increasing feature channels from 32 to 256 while reducing spatial resolution. The decoder mirrors this hierarchy, incorporating scSE blocks, skip-connection concatenation, and bilinear upsampling to recover fine spatial details, followed by a $3 \times 3$ convolution with sigmoid activation to generate the normalized output. The loss function combines mean squared error (MSE) and structural similarity index measure (SSIM)[24], balancing pixel-wise accuracy and structural preservation to avoid over-smoothing and enhance reconstruction of fine vascular textures.

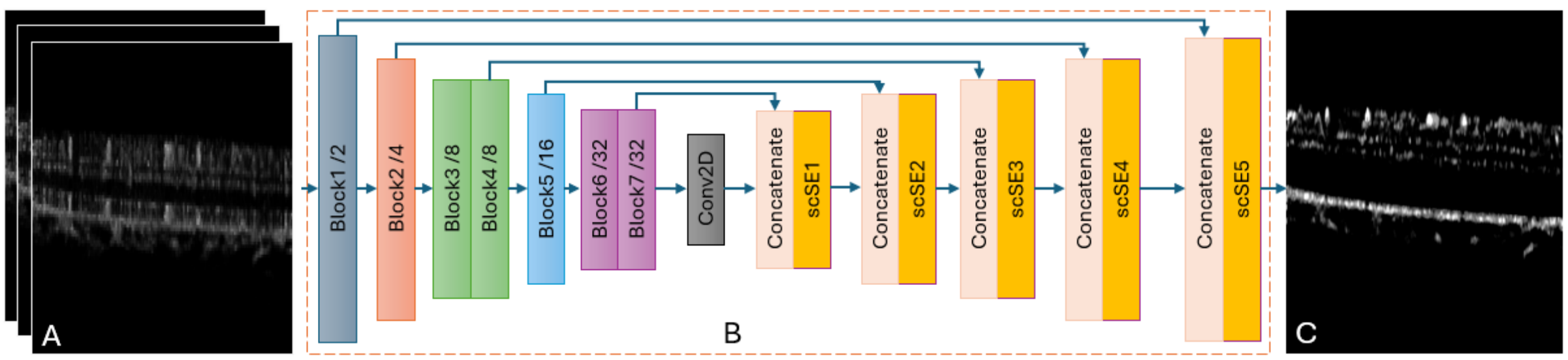

Figure 1. Architecture of the proposed deep learning model for OCTA signal restoration. (A) Input: three adjacent OCTA B-frames. (B)Encoder: EfficientNet-B5; decoder: integration of spatial and channel squeeze-and-excitation (scSE) modules. (C) Output: cleaned middle OCTA B-frame.

***Ground Truth Construction***

Artifact-free OCTA volume was synthesized from the multiple repeated acquisitions. For each eye, 6–9 OCTA volumes were available in the dataset. The ground-truth generation pipeline consisted of three steps. First, projection artifacts were removed from each individual volume using our previously published OCTA projection artifact removal algorithm [17]. Second, the artifact-suppressed volumes from the same eye were registered at the A-line level to correct sub-voxel motion, as demonstrated in our prior work [25]. Third, the aligned volumes were averaged voxel-wise to generate a high signal-to-noise ratio (SNR) volume. For each eye, one original single OCTA volume was randomly selected from the repeated scans as the network input [Fig. 2A], while the averaged volume served as the ground truth [Fig. 2B].

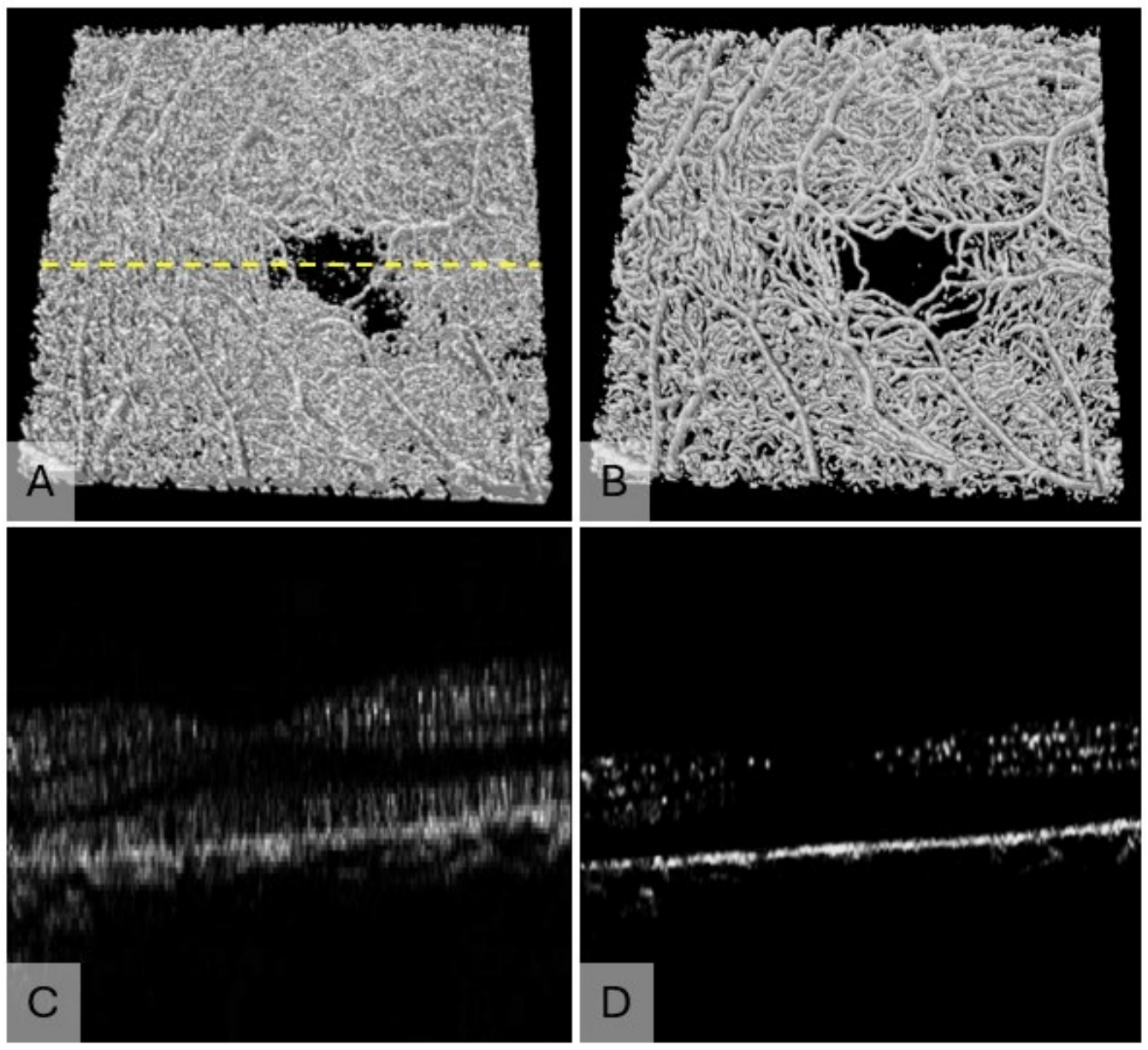

Figure 2. Comparison of original single OCTA volume and registered multi-volume OCTA volume. (A) Original single OCTA volume. (B) Registered multi-volume OCTA volume. (C) Cross-sectional OCTA B-frame from the original volume at the location indicated by the yellow dashed line in (A). (D) Corresponding cross-sectional OCTA B-frame from the ground-truth volume.

**Training Configuration and Hyperparameters**

***Loss function and optimizer***

The network was trained end-to-end by minimizing a composite loss that combines mean squared error (MSE) and the structural similarity index measure (SSIM) loss with equal weighting:

$$\mathrm{MSE} = \frac{1}{N}\sum(y - \hat{y})^2 \tag{1}$$

$$\mathrm{SSIM} = \frac{\left(2 \cdot \mu_y \mu_{\hat{y}} + C_1\right)\left(2 \cdot \sigma_{y\hat{y}} + C_2\right)}{\left(\mu_y^2 + \mu_{\hat{y}}^2 + C_1\right) \cdot \left(\sigma_y^2 + \sigma_{\hat{y}}^2 + C_2\right)} \tag{2}$$

$$\mathrm{Loss} = \mathrm{MSE} + (1 - \mathrm{SSIM}) \tag{3}$$

where $\mu_y$ and $\mu_{\hat{y}}$ are the local mean intensities of the ground truth $y$ and model output $\hat{y}$ , $\sigma_y^2$ and $\sigma_{\hat{y}}^2$ are the local variances, $\sigma_{y\hat{y}}$ is the local covariance, and $C_1 = (k_1 L)^2$, $C_2 = (k_2 L)^2$ are numerical stability constants with $k_1 = 0.01$, $k_2 = 0.03$, and $L = 1.0$ (normalized dynamic range).

The MSE term enforces pixel-wise fidelity by penalizing large intensity deviations, ensuring accurate restoration of mean vascular signal. The SSIM term complements this by optimizing perceptual quality through intensity, contrast, and structural similarity, thereby preserving fine capillary morphology that may be oversmoothed by MSE alone. Together, these losses provide complementary constraints: MSE stabilizes training and limits local errors, while SSIM promotes structural coherence at the vessel level.

The optimizer used the Adam algorithm [26] with an initial learning rate of 0.001. A reduce-on-plateau strategy decreased the learning rate by a factor of 0.1 if validation loss did not decrease for 3 consecutive epochs. Training was conducted for up to 500 epochs with early stopping (patience = 7), and the model with the lowest validation loss was retained for evaluation. All convolution layers applied L2 weight regularization.

### *Data Augmentation*

All volumetric OCTA data were normalized to [0, 1] using min-max scaling. Data augmentation operations were applied to training datasets during training, including random scale-and-crop, random brightness jitter, adding Gaussian noise and random horizontal flip.

### Evaluation

Model performance was evaluated using Peak signal-to-noise ratio (PSNR) and SSIM (Eq. 2). PSNR measures the ratio of the maximum signal power to the mean squared reconstruction error

$$\mathrm{PSNR} = 10 \cdot \log_{10}\left(\frac{1}{\mathrm{MSE}}\right) \tag{4}$$

where intensities are normalized to [0, 1]. MSE, defined in Eq. 1, quantifies the mean per-pixel squared error between the model output and the ground truth. Higher PSNR indicates smaller absolute pixel-level deviations from the reference, effectively capturing global errors like spurious background noise. Five-fold cross-evaluation was performed in the performance evaluation.

To evaluate model performance in microvasculature restoration, we computed and compared the microvascular skeleton consistency between the original OCTA volumes and the ground truth, as well as between the model outputs and the ground truth. Since skeletonization is highly sensitive to the binarization of OCTA images, which means even single-pixel offsets can cause complete mismatches in the resulting skeleton. Thus, exact pixel-level comparison is an unreliable measure of the overlap of skeletonized microvasculature. To address this, we introduce skeleton- tolerant Dice Coefficient (skDice) as defined in Eq.5, which relaxes the strict overlap requirement of the standard Dice Coefficient: a pixel (or voxel in 3D) is considered ‘matched’ if it falls within a morphological dilation of the reference skeleton, rather than requiring exact co-location.

$$\mathrm{skDice} = \frac{\mathrm{Dilation}(y) \cdot \hat{y} + \mathrm{Dilation}(\hat{y}) \cdot y}{y + \hat{y}} \tag{5}$$

where $y$ is the skeletonized ground truth and $\hat{y}$ is the skeletonized model output. The Dilation operator performs morphological dilation using a square structuring element (cubic in 3D) of size 3. The evaluation was performed on both 2D *en face* OCTA images and 3D OCTA volumes. The 3D volumes included slabs from the superficial vascular complex (SVC), deep vascular complex (DVC), and the whole inner retina. The 2D *en face* images were generated by projecting the 3D slabs onto 2D planes using a maximum value projection method [27]. The microvasculature was measured in skeletonized.

To further evaluate the performance of proposed model in the clinical vasculature biomarker quantification. We measured the nonperfusion area (NPA) in the SVC, DVC, and whole inner retina use our previous proposed AI model [10]. Repeatability of NPA measurements was evaluated in 71 eyes with repeated scans from an independent dataset. Each eye included at least two repeated scans, yielding a total of 423 scans. The dataset comprised eyes with vision-threatening diabetic retinopathy, defined as ETDRS level ≥ 53.

## Results

The model was trained on 120 OCTA volumes and evaluated on 12,000 B-frame drawn from the test set (30 volumes × 400 B-scan per volume). The additional dataset with 71 eyes was used to assess the model’s repeatability.

### Performance evaluation of OCTA signal restoration

The model performance was evaluated using PSNR and SSIM (Table 1). The restored angiograms significantly improved the PSNR by 17.7% (p<0.001) and SSIM with improvement 26.7% (p<0.001).

**Table 1. Performance of model on OCTA restoration (mean ± standard deviation)**

| Metric | Original OCTA | Restored OCTA | Improvement (%) | *p*-value |
|---|---|---|---|---|
| **PSNR (dB)** | 22.23±0.78 | 26.16±1.26 | +17.7% | <0.001 |
| **SSIM** | 0.72 ± 0.03 | 0.91±0.02 | +26.7% | <0.001 |

Compared with the original single OCTA volume [Fig. 3A], the restored OCTA [Fig. 3B] exhibits reduced background noise and clearer retinal vasculature, closely matching the ground truth [Fig. 3C]. This improvement is also evident in the cross-sectional B-frames, where the original images [Fig. 3D] show fragmented and noisy vascular signals, whereas the restored B-frame [Fig. 3E] reveals continuous, well-defined vessels with enhanced contrast, approaching the ground truth appearance [Fig. 3F].

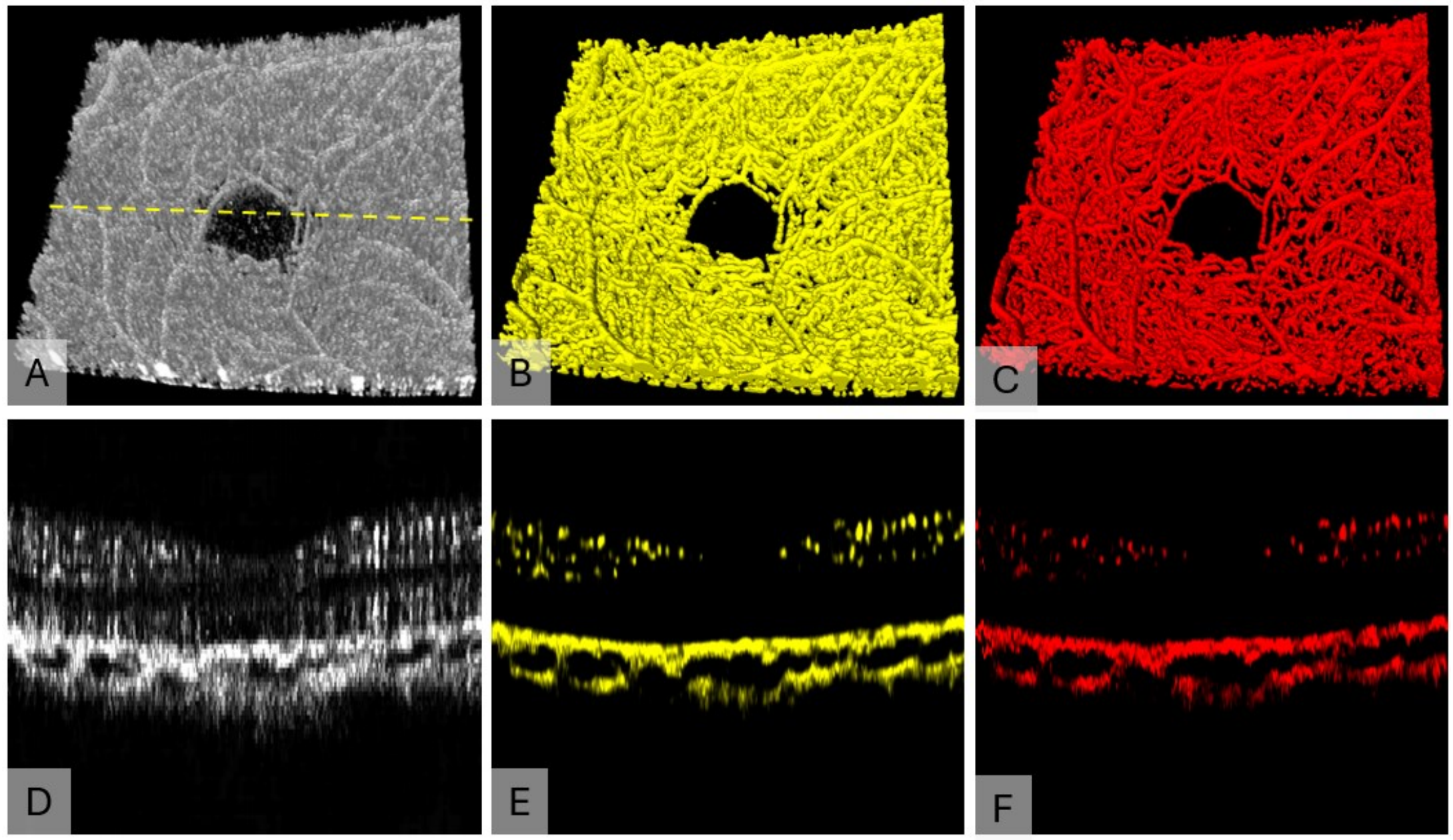

Figure 3. Example of OCTA restoration. (A) The original single OCTA. (B) OCTA restored by the proposed deep learning model. (C) Ground-truth OCTA. (D–F) Corresponding cross-sectional OCTA images at the location marked in (A): (D) original single OCTA, (E) restored OCTA, and (F) ground truth. The volume rendering (A-C) only includes retina for better visualization. The Cross-sectional images include both retina and choroid.

**Microvascular Skeleton Consistency**

Microvascular morphology was quantified by extracting binary skeletons from the 2D *en face* projections and 3D volumes from slabs of the SVC, DVC, and the whole inner retina [Fig. 4], then computing the skDice between the restored (or original) skeletons and the ground-truth skeletons.

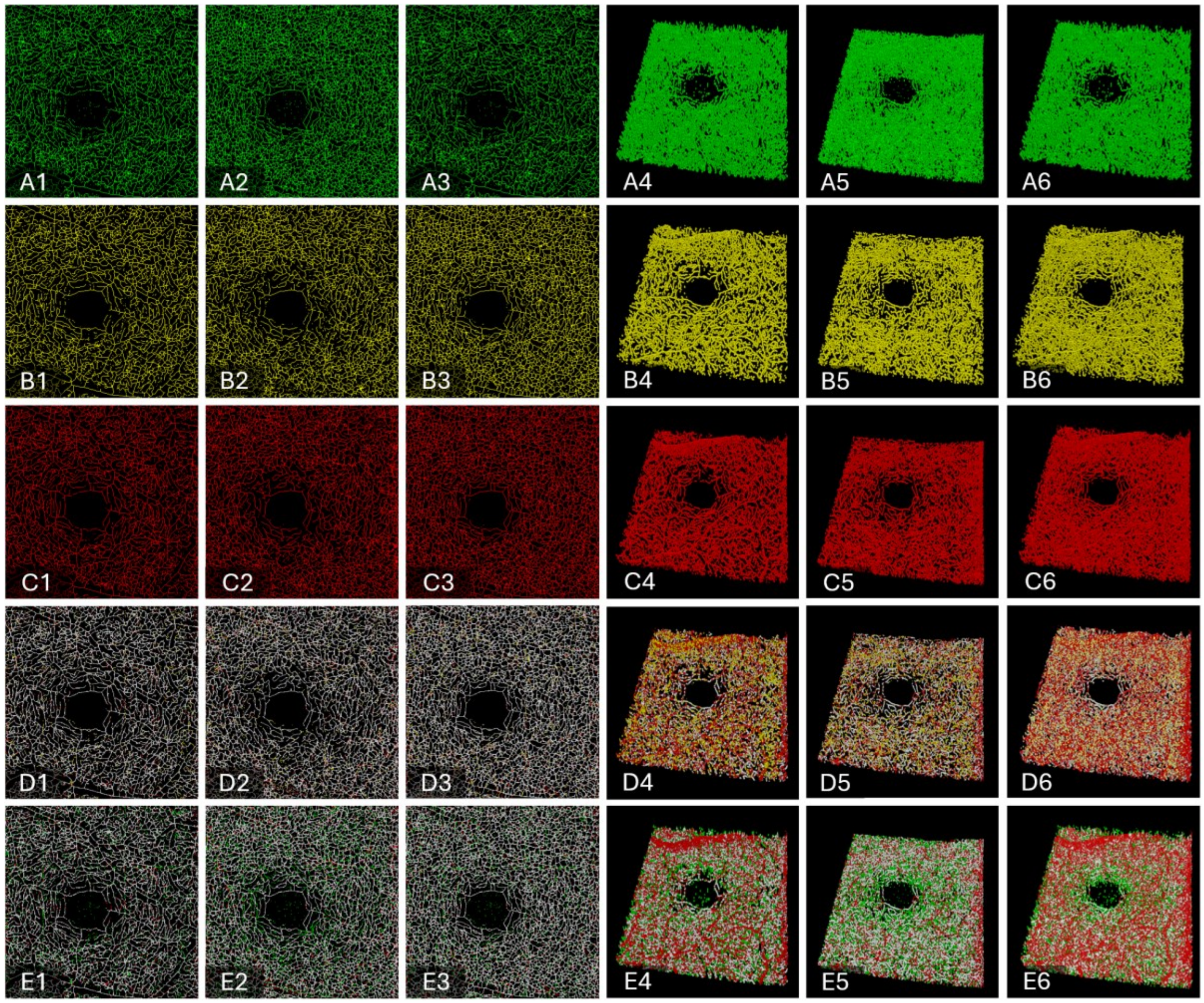

Figure 4. Examples of skeletonized OCTA generated from original OCTA volumes, ground truth, and model outputs. A1–A6: Skeletonized 2D en face and 3D OCTA volumes of the SVC, DVC, and whole inner slab derived from the original OCTA volumes. B1–B6: Skeletonized 2D en face and 3D OCTA volumes of the SVC, DVC, and whole inner slab derived from the model output OCTA volumes. C1–C6: Skeletonized 2D en face and 3D OCTA volumes of the SVC, DVC, and whole inner slab derived from the ground truth OCTA volumes. D1–D6: Overlaid skeletonized 2D en face and 3D OCTA volumes of the SVC, DVC, and whole inner slab comparing model outputs with ground truth (white: overlap; yellow: model output; red: ground truth). E1–E6: Overlaid skeletonized 2D en face and 3D OCTA volumes of the SVC, DVC, and whole inner slab comparing original OCTA with ground truth (white: overlap; green: original OCTA; red: ground truth).

In each of the regions considered, the model produced significantly better skDice scores when compared to the original angiograms [Figure 5]. The difference is especially striking in the 3D restoration, where the distributions only barely overlapped. These gains show that restoring consistency between slices helps recover the 3D vessel network that is mostly missing in the original data.

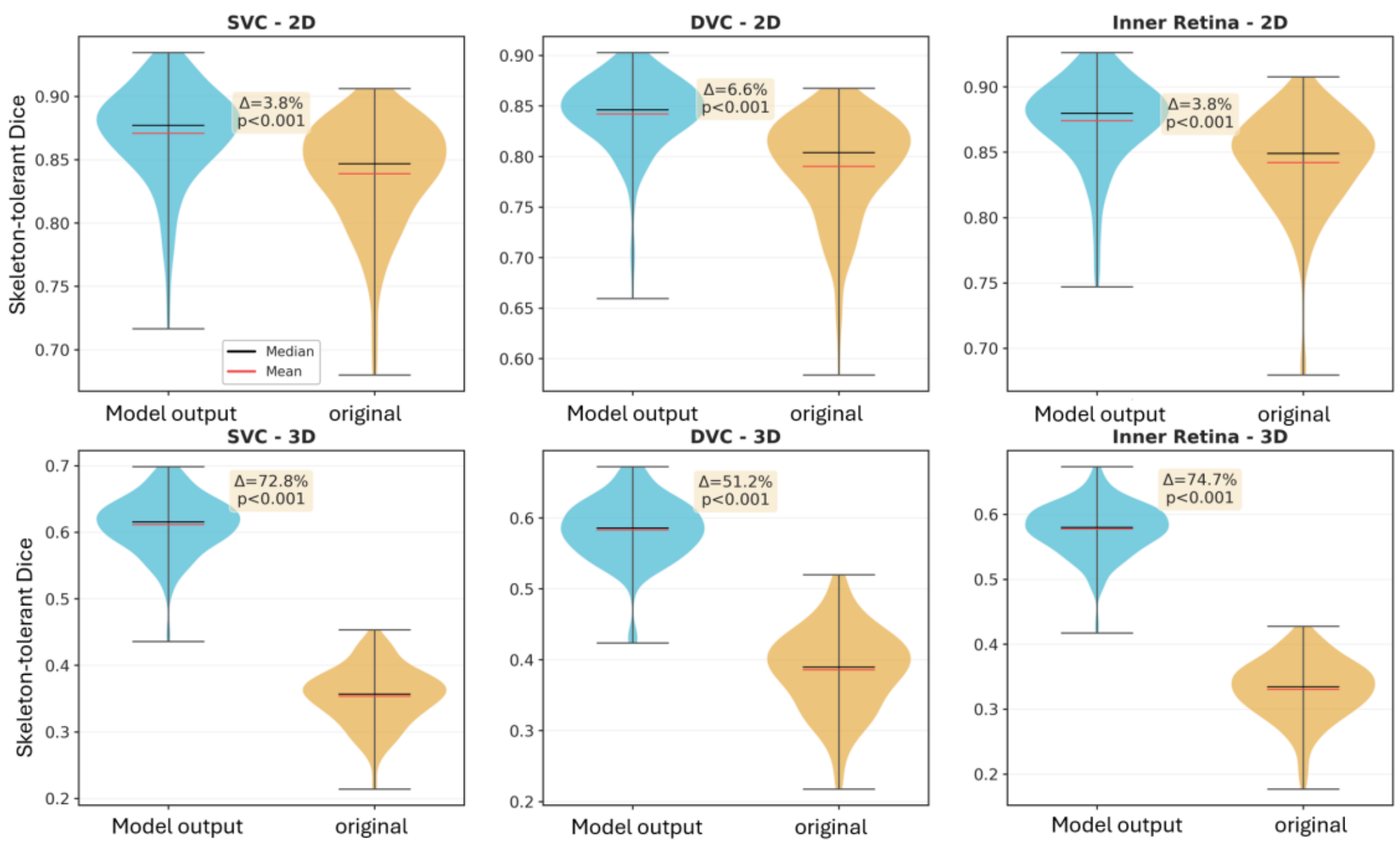

Figure 5. Skeleton-tolerant Dice Coefficients for the SVC, DVC, and inner retina in 2D (top) and 3D (bottom) comparing restored OCTA (blue) and original single OCTA (orange) against the ground truth.

**Non-Perfusion Area Repeatability**

NPA was quantified using our previously published AI model [10] across the SVC, DVC, and whole inner retina [Figure 6]. The restoration significantly improved the repeatability of NPA measurements in the SVC and inner retina (both $p < 0.001$). In the DVC, repeatability showed a slight non-significant improvement ($p = 0.459$) [Table 2].

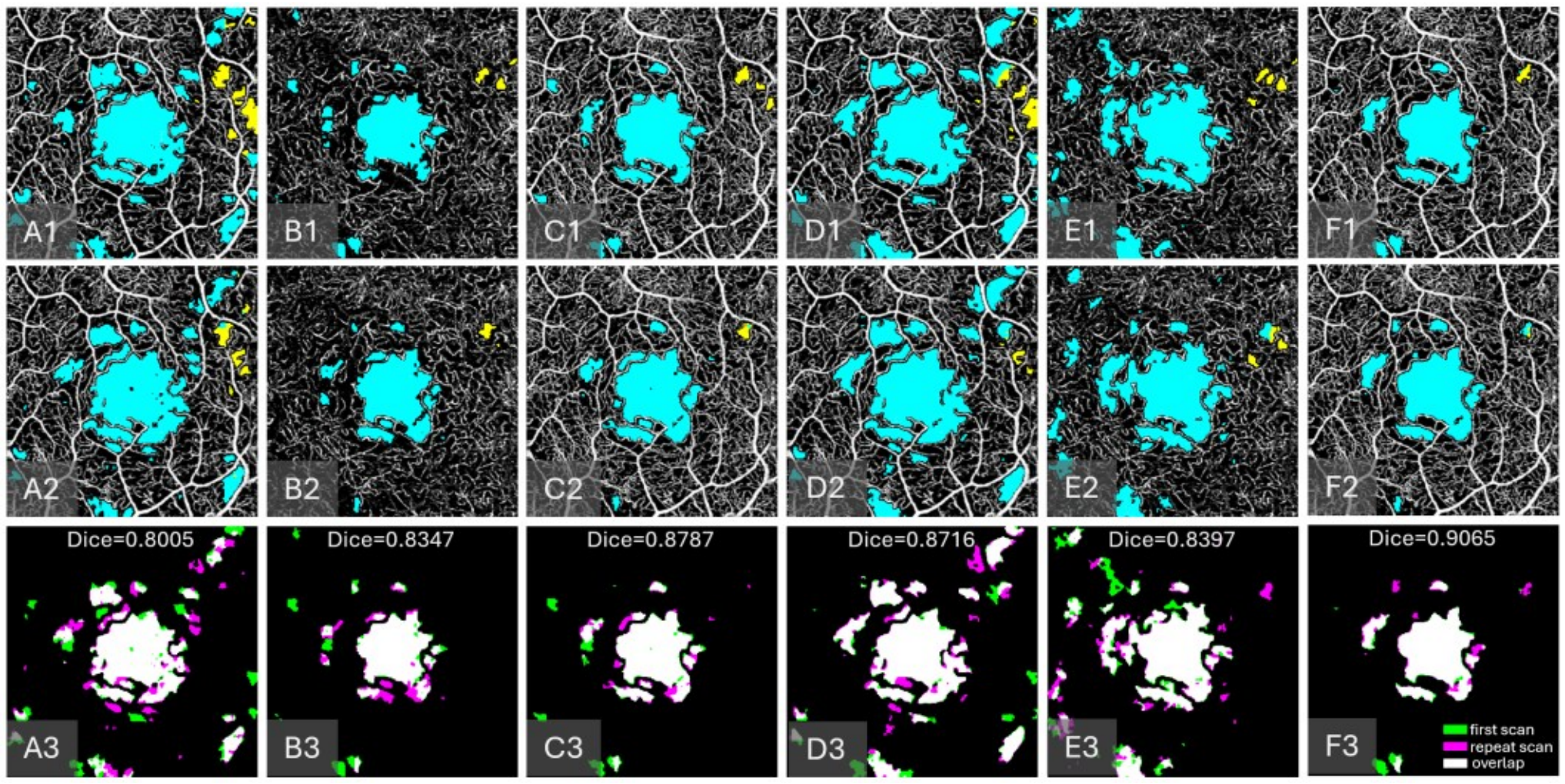

Figure 6. The nonperfusion area measurements on repeated OCTA volumes. A1-C1, the NPA overlayed on *en face* OCTA of SVC, DVC and whole inner from the first original OCTA volume, (blue: NPA, yellow: shadow artifacts). A2-C2, the NPA overlayed on *en face* OCTA of SVC, DVC and whole inner from repeated original OCTA volume. A3-C3, the overlapping NPA of original OCTA volume. D1-F1, the NPA overlayed on *en face* OCTA of SVC, DVC and whole inner from the first model output OCTA volume. D2-F2, the NPA overlayed on *en face* OCTA of SVC, DVC and whole inner from repeated model output OCTA volume. A3-F3, the overlapped NPA of model output OCTA volume.

**Table 2. Repeatability of non-perfusion area measurements for restored and original OCTA images (CV*, mean ± standard deviation)**

| Slab | Original OCTA (%) | Restored OCTA (%) | p-value |
|---|---|---|---|
| **SVC** | 8.83 ± 6.59 | 5.71 ± 4.93 | <0.001 |
| **DVC** | 10.75 ± 9.22 | 10.24 ± 7.69 | 0.459 |
| **Inner Retina** | 9.32 ± 7.73 | 5.23 ± 4.72 | <0.001 |

*CV (Correlation Variance) $= 100 \cdot \mu/\sigma$ , where $\mu$ and $\sigma$ are mean and standard deviation of NPA area between two repeats

## Discussion

In this study, we developed a deep learning model to enhance and restore 3D retinal microvasculature from single OCTA volumes. The method reconstructs high–quality 3D vascular structures by simultaneously suppressing multiple types of artifacts while preserving capillary morphology. Unlike prior approaches that operate primarily on 2D *en face* projections or target individual artifacts, our method addresses all types of artifacts directly in the volumetric domain. By leveraging the retinal vasculature's strong structural continuity, the model reconstructs 3D vasculature from a single noisy acquisition, reducing the need for repeated scans or extensive post-processing.

To overcome limitations of previous approaches to improve OCTA image quality, our framework introduces several key innovations. First, the model operates directly on OCTA B-frames in the cross-sectional volumetric domain, rather than on 2D *en face* projections. By processing three consecutive B-frames as a stacked input and predicting the restored middle frame, the model exploits inter-slice structural continuity along the B-scan axis to enforce 3D vascular coherence, which prior 2D approaches may miss. This architecture enables simultaneous suppression of multiple artifact types within a single model, without requiring separate processing steps for each. Second, the EfficientNet-B5 encoder, combined with scSE modules in the decoder, enables selective, multi-scale feature extraction across both spatial and channel dimensions, preserving fine capillary structures that would otherwise be oversmoothed by a simpler architecture. Third, the composite MSE and SSIM loss function is specifically designed to balance pixel-level intensity fidelity with structural perceptual quality. Finally, our ground-truth construction pipeline, which integrated projection artifact removal, multi-scan registration, and volume averaging to generate artifact-free ground truth that provides reliable training data for the proposed model.

Our quantitative evaluation demonstrates substantial improvements in OCTA signal quality. The proposed model increased the PSNR by approximately 4 dB and improved the SSIM by nearly 27% compared with the original single OCTA volumes [Table 1]. These improvements indicate that the network effectively suppresses artifacts while maintaining vascular morphology [Fig. 3]. Because multi-scan acquisition increases imaging time and is susceptible to motion artifacts, the ability to approximate multi-scan image quality with a single acquisition may offer practical benefits for clinical OCTA workflows. More importantly, the microvascular skeleton Dice overlaps between restored volumes and ground truth [Fig. 4] improved significantly across all evaluated slabs, with 3D Dice scores approximately doubling relative to the original OCTA [Fig. 5]. This improvement in 3D skeletal consistency reveals a key finding: artifacts in original OCTA systematically disrupt signal throughout the volumetric OCTA. Beyond improvements in signal quality metrics, our framework also significantly enhanced the repeatability of clinically relevant quantitative biomarkers [Fig. 6]. Repeat-scan analysis demonstrated a significant reduction in the coefficient of variation for NPA measurements in both the SVC (8.83% to 5.71%, $p < 0.001$) and the whole inner retinal slab (9.32% to 5.23%, $p < 0.001$) [Table 2]. The non-significant improvement in the DVC ($p = 0.459$) likely indicates that the DVC is more vulnerable to signal-attenuation artifacts and noise. Future approaches that incorporate structural OCT or physics-informed attenuation modeling may further improve DVC restoration.

Several limitations of our work should be noted. First, the training data set was acquired using a single OCTA system and scanning protocol. While the model performed well within this dataset, its generalizability to other devices or acquisition settings remains untested. Second, the ground-truth volumes were generated via multi-scan averaging after projection artifact removal, which may retain residual artifacts or imperfect registration. Third, although our method improves image quality and quantitative repeatability, it does not explicitly incorporate physical models of OCTA signal formation or projection artifact mechanisms. Instead, the model estimates the true OCTA signal directly from the original single OCTA data. Finally, although the dataset included eyes with multiple retinal diseases, validation in larger and more diverse clinical cohorts will be necessary to confirm the robustness of the approach.

Future work will focus on enhancing the robustness and clinical applicability of the proposed framework. Validation on multi-center datasets acquired with different OCTA systems will be necessary to assess cross-device generalization. Incorporating physics-informed learning or explicit modeling of projection artifacts may further improve restoration accuracy in deeper retinal layers. In addition, integrating structural OCT information may also help the network better distinguish true blood flow signal from projection artifacts. Ultimately, embedding the restoration algorithm within real-time OCTA acquisition pipelines may enable on-the-fly enhancement of OCTA volumes, improving both visualization and quantitative analysis during routine clinical imaging.

## Conclusion

This study demonstrates that deep learning–based volumetric restoration can restore the OCTA signal from the original OCTA volume, enhance the reconstruction of 3D retinal microvasculature, and improve the repeatability of clinically relevant vascular biomarkers. By enabling reliable visualization and quantification of retinal microvasculature from a single OCTA, the proposed framework may facilitate more accurate clinical interpretation and automated analysis of OCTA data.

**Funding:** This work was supported by grants from National Institutes of Health (R01 EY 036429, R01 EY035410, R01 EY024544, R01 EY027833, R01 EY031394, R43EY036781, P30 EY010572, T32 EY023211, UL1TR002369); the Jennie P. Weeks Endowed Fund; the Malcolm M. Marquis, MD Endowed Fund for Innovation; Unrestricted Departmental Funding Grant and Dr. H. James and Carole Free Catalyst Award from Research to Prevent Blindness (New York, NY); Edward N. & Della L. Thome Memorial Foundation Award, and the Bright Focus Foundation (G2020168, M20230081).

**Financial interests:** Yukun Guo: Optovue/Visionix (P), Genentech/Roche (P, R); Ifocus Imaging (P); Tristan.T. Hormel: Ifocus Imaging (I); Steven T. Bailey: Optovue/Visionix (F); Yali Jia: Optovue/Visionix (P, R), Genentech/Roche (P, R, F), Ifocus Imaging (I, P), Optos (P), Boeringer Ingelheim (C), Kugler (R)